\documentclass[conference]{IEEEtran}
\IEEEoverridecommandlockouts
\usepackage{cite}
\usepackage{amsmath,amssymb,amsfonts}
\usepackage{cleveref}
\usepackage{algorithm}
\usepackage{algorithmic}
\usepackage{graphicx}
\usepackage{subcaption}
\usepackage{textcomp}
\usepackage{xcolor}
\usepackage{dsfont}
\usepackage{array,multirow}
\usepackage{tabularx,booktabs}
\newcolumntype{C}{>{\centering\arraybackslash}X} 
\DeclareUnicodeCharacter{2061}{X}
\usepackage{adjustbox}
\usepackage{dashbox}
\def\BibTeX{{\rm B\kern-.05em{\sc i\kern-.025em b}\kern-.08em
    T\kern-.1667em\lower.7ex\hbox{E}\kern-.125emX}}
\begin{document}

\title{Multi-Grid Redundant Bounding Box Annotation for Accurate Object Detection \\
}

\author{\IEEEauthorblockN{Solomon Negussie Tesema}
\IEEEauthorblockA{\textit{ImViA Laboratory} \\
\textit{University of Burgundy}\\
Dijon, France \\
solomon.negussie.tesema@gmail.com}
\and
\IEEEauthorblockN{El-Bay Bourennane}
\IEEEauthorblockA{\textit{ImViA Laboratory} \\
\textit{University of Burgundy}\\
Dijon, France\\
ebourenn@u-bourgogne.fr}
}

\maketitle 

\begin{abstract}
Modern leading object detectors are either two-stage or one-stage networks repurposed from a deep CNN-based backbone classifier network. YOLOv3 is one such very-well known state-of-the-art one-shot detector that takes in an input image and divides it into an equal-sized grid matrix. The grid cell having the center of an object is the one responsible for detecting the particular object. This paper presents a new mathematical approach that assigns multiple grids per object for accurately tight-fit bounding box prediction. We also propose an effective offline copy-paste data augmentation for object detection. Our proposed method significantly outperforms some current state-of-the-art object detectors with a prospect for further better performance. 
\end{abstract}

\begin{IEEEkeywords}
Object detection, Multi-grid assignment, Copy-Paste Image augmentation
\end{IEEEkeywords}

\section{Introduction}
An object detection network aims to locate an object on an image using a tight-fit rectangular bounding box and label it correctly. Nowadays, there are two distinct approaches to achieve this purpose. The first and performance-wise, the most dominant approach is two-stage object detection, best represented RCNN \cite{rcnn} and its derivatives \cite{fasterrcnn, fastrcnn}. In contrast, the second set of object detection implementations, well acknowledged for their outstanding detection speed and light-weightiness, are referred to as one-staged networks, representative examples being \cite{yolov1}, \cite{ssd},\cite{retinanet}. Two-stage networks rely on an underlying region proposal network that generates candidate regions of an image likely to contain an object of interest, and a second detection head handles the classification and bounding box regression. In one-stage object detection, detection is a single, fully unified regression problem that simultaneously handles classification and localization in one complete forward pass. Due to this, usually, one-stage networks are lighter, faster, and simple to implement. 

One-stage networks can be classified as anchor-based \cite{yolov1, yolov2,yolov3,yolov4, ssd, retinanet, denseyolo, ddgnet}  and anchorless \cite{centernet, cornernet}. Anchor-based networks such as YOLOv3\cite{yolov3} or YOLOv4\cite{yolov4} mainly divide an input image into equal grid cells. Furthermore, each grid cell regresses object bounding box coordinates and classifies them into one of the predefined class categories while simultaneously scoring the predicted box’s objectness confidence. Other anchor-based variants, such as SSD \cite{ssd} and RetinaNet \cite{retinanet} employ the concept of feature pyramid to perform multi-scale detection extracted from different layers of the backbone classifier network such as VGG \cite{vgg} or families of ResNet \cite{resnet}. Recent anchorless entries of one-shot detectors use key-point such as corners of bounding box \cite{cornernet} or center coordinate or combination of both \cite{centernet}, replacing the use of anchor boxes by key-point pooling convolutional pipelines along their bounding box regression and classification networks. 

This paper sticks to YOLO’s approach, particularly YOLOv3 \cite{yolov3}, and proposes a simple hack that simultaneously enables multiple grid cells to predict an object coordinate, class, and objectness confidence. The basic theory behind multi-grid cell assignment per object is to increase the likelihood of predicting a tight-fit bounding box by enforcing more than one cell working on the same object. Some of the advantages of multi-grid assignment includes: (a) gives the object detector a multi-perspective view of the object it is detecting rather than relying on just one grid cell to predict the class and the coordinates of an object, (b) less random and erratic bounding box prediction, meaning high precision and recall, since nearby grid cells are trained to predict same object class and coordinates,(c) reducing the imbalance between grid cells with an object of interest against grids without an object of interest. Moreover, since the multi-grid assignment is mathematical utilization of an existing parameters and does not require an extra keypoints pooling layer and post-processing to regroup keypoints to their corresponding objects like CenterNet\cite{centernet} and CornerNet\cite{cornernet}, we say it is a more natural way of achieving what anchorless or keypoint-based object detectors are trying to achieve. In addition to the multi-grid redundant annotation, we also introduce a new offline copy-paste-based data augmentation technique for accurate object detection. 
 
\section{Related Works}
The pioneer and most successful one stage-detector YOLO \cite{yolov1} and its successive incremental improvements \cite{yolov2}, \cite{yolov3}, \cite{yolov4} divide an input image into grid cells of equal size. The grid that contains the center of a given object-bounding box on an image is responsible for detecting that particular object. Since YOLOv1, the authors of YOLO tried to improve the performance of their object detector by incrementally incorporating key improvements such as more network depth, more anchor boxes, slight change on loss function, and lately since YOLOv3 best practices such as multi-scale detection and skip-connections are incorporated.  

The other typical one-stage detector is SSD\cite{ssd}. SSD uses multi-layer feature pyramids on top of a backbone classification network, notably a VGG network,  to perform a multi-scale detector that better handles objects of various scales. Using a similar concept of feature pyramids as in SSD, another famous object detector called RetinaNet \cite{retinanet} proposed a novel loss function called focal-loss to solve the foreground and background class imbalance prevalent in one-stage detectors unlike two-stage networks.  

Recently, anchorless one-stage object detection techniques such as \cite{cornernet}, \cite{centernet}, \cite{fcos}, \cite{densebox} aim to reduce the hurdle of determining the appropriate number and shape of anchor boxes. Networks such as DSSD \cite{fu2017dssd} and RetinaNet \cite{retinanet} use default-boxes, also referred to as anchor boxes,  amounting over tens or hundreds of thousands, resulting in slow training and brutal non-max suppression during inference. Instead, anchorless detectors add a separate layer to pool and process points on the bounding box of an image. CornerNet\cite{cornernet}, for example, adds a pipeline that processes the corner keypoints of an object hence needing no anchor boxes. CenterNet\cite{centernet}, another anchorless one-stage detector, adds a third point, namely the center point of an object bounding box in addition to the corner keypoints. The center keypoint in CenterNet is to aid CornerNet to have a more global view of an object it is detecting, which was its bottleneck at first. 

This paper sticks to YOLO’s grid-based approach since YOLO’s approach neither requires many anchor boxes like SSD, DSSD or RetinaNet nor adds a separate pipeline to process keypoints like CornerNet or CenterNet. However, unlike YOLO, we propose a mathematical way to assign an object to multiple grid cells, including the grid cell where the center of the object-bound box falls. As we stated earlier, in YOLO, the grid that contains an object’s bounding box center is made responsible for detecting that particular object, hence, one grid assignment per object. In our implementation, we will show that mathematically it is possible to assign any number of grid cells to annotate an object, though we will only use the grids around the center, including the center grid. Due to the multi-grid annotation, we dubbed our object detector MultiGridDet short for Multi-Grid Detector. Our detector is light and faster than YOLOv3 mainly due to two reasons; (1) MultiGridDet has relatively less depth number of layers and (2) we use a lighter output layer, or detection head, based on the technique introduced by DenseYOLO \cite{denseyolo}. 

\section{Multi-Grid Assignment}
\begin{figure*}[h]
\centerline{\includegraphics[width=0.75\textwidth]{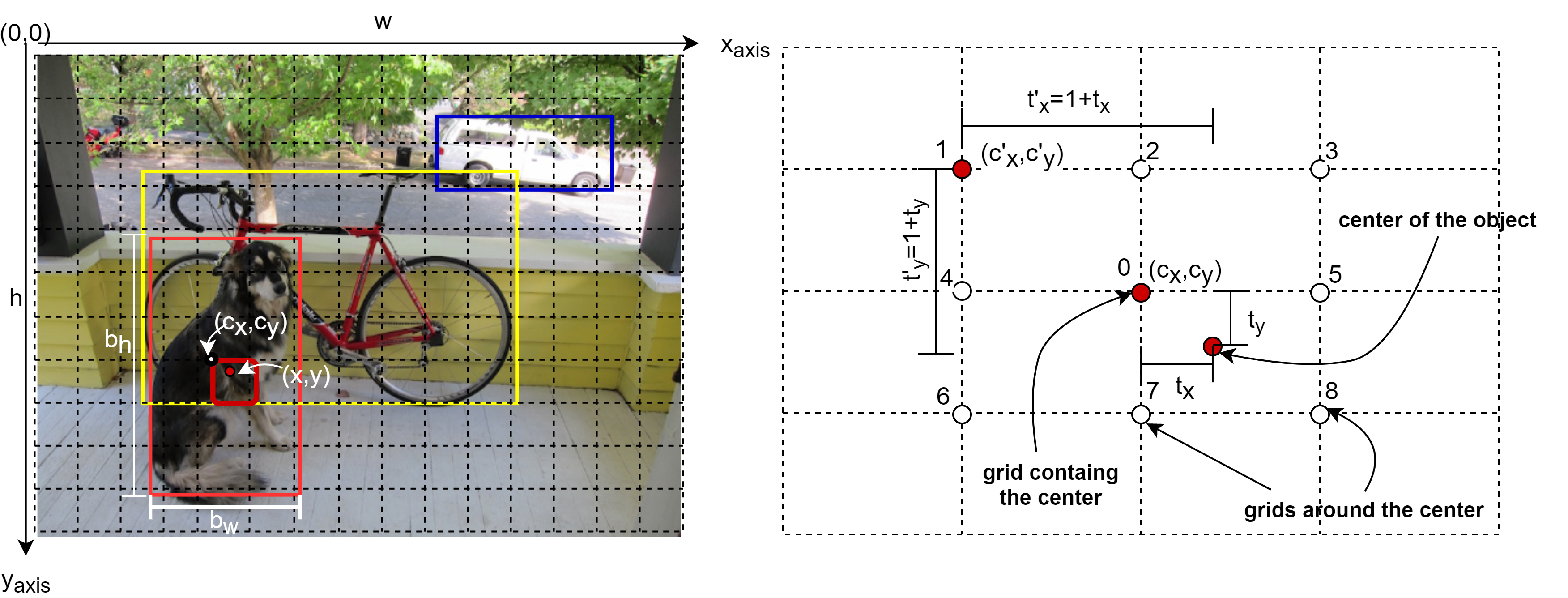}}
\caption{\textbf{Multi-grid assignment}}
\label{multigridassignment}
\end{figure*}

Consider Figure \ref{multigridassignment} containing three objects, namely a dog, bicycle, and car. For brevity, we will explain our muti-grid assignment on one object, the dog. Figure \ref{multigridassignment}(a) shows the three objects bounding box with more detail on the dog’s bounding box. Figure \ref{multigridassignment}(b) shows the zoomed-out region of Figure \ref{multigridassignment}(a), focusing on the dog’s bounding box center. The top-left coordinate of the grid cell containing the center of the dog bounding box is labeled by number 0, while the other eight grid cells around the grid containing the center have a label from 1 up to 8. 

In YOLO and other YOLO-based detection networks, the grid labeled 0 is solely responsible for predicting the class dog and its precise bounding box coordinates $(x,y, b_w, b_h)$, whereas in our case, we assign all grids labeled 0 to 8 to predict the class and the precise bounding box of the dog simultaneously.

The grid containing the center coordinate, the small red box on Figure \ref{multigridassignment}(a), or coordinate labeled 0 on Figure \ref{multigridassignment}(b) is at the grid location $\left(c_x,\ c_y\right)$. We calculate the $\left(c_x,\ c_y\right)$ coordinates using the equation, $c_x\ =\ \left\lceil\frac{x}{g_w}\right\rceil$ and $c_y\ =\ \left\lceil\frac{y}{g_h}\right\rceil$, where $g_w$ and $g_h$ are the grid cell’s width and height, respectively. In YOLOv1 and YOLOv2 $g_w$ and $g_h$ are both 32 pixels each, whereas in YOLOv3, due to the multi-scale detection feature tailored for small, medium and large scale object, the grid cells are also in those three scales; $8\times 8$, $16 \times 16$, and $32 \times 32$ pixels. 

Starting from YOLOv2, YOLO-based object detectors predict an offset of the bounding box from pre-generated anchor boxes instead of directly predicting the bounding box’s coordinates. As a result, the ground-truth bounding box values, $(x,y)$ and $(b_w, b_h)$ are rescaled to smaller scales $(t_x, t_y)$ and $(t_w, t_h)$, respectively, for training stability using \Crefrange{equ:one}{equ:four}. Note that, $t_x$ and $t_y$ are in the range $[0, 1]$. 

\begin{equation}\label{equ:one}
	t_x=\frac{x}{g_w}\ -\ \left\lceil\frac{x}{g_w}\ \right\rceil\ \end{equation}
\begin{equation}\label{equ:two}
	t_y=\frac{y}{g_h}\ -\ \left\lceil\frac{y}{g_h}\ \right\rceil
\end{equation}
\begin{equation}\label{equ:three}
	t_w=\log{\left(\frac{b_w}{a_w}\right)}
\end{equation}
\begin{equation}\label{equ:four}
	t_h=\log{\left(\frac{b_h}{a_h}\right)}
\end{equation}, where $a_w$, $a_h$ are the best-fit anchor box's width and height, respectively, generated using K-means IoU clustering. 

One can easily convert the $(t_x, t_y)$ and $(t_w, t_h)$ parameters to the original bounding box parameters of an object, that is $(x,y)$ and $(b_w, b_h)$ using the reverse \Crefrange{equ:five}{equ:eight}:
\begin{equation}\label{equ:five}
	x\ =\ \left(c_x\ +\ t_x\right)\times g_w
\end{equation}
\begin{equation}\label{equ:six}
	y\ =\ \left(c_y\ +\ t_y\right)\times g_h
\end{equation}
\begin{equation}\label{equ:seven}
	b_w\ =\ a_w\times e^{t_w}
\end{equation}
\begin{equation}\label{equ:eight}
	b_h\ =\ a_h\times e^{t_h}
\end{equation}

\begin{figure*}[h]
\centerline{\includegraphics[width=0.75\textwidth]{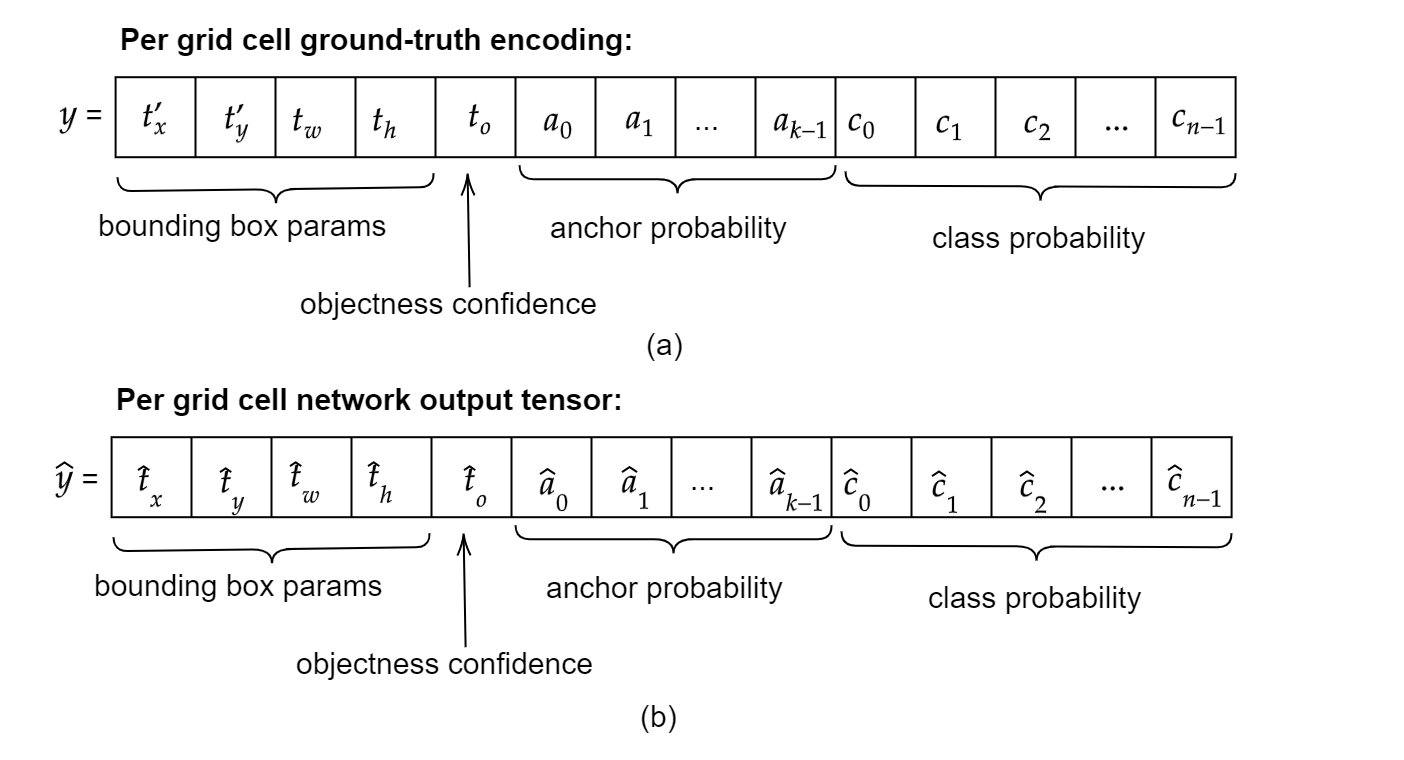}}
\caption{\textbf{Ground-truth encoding}}
\label{objectencoding}
\end{figure*}

Thus far, we have explained how the grid containing the center of an object’s bounding box annotates an object’s ground truth. This dependence on just one grid cell per object to do the difficult job of predicting the class and the exact tight-fit bounding box raises many questions such as (a) massive imbalance between the positive and negative grids, that is, grids with and without object’s center coordinate  (b)  slow bounding box convergence to ground-truth, (c) lack of multi-perspective (angle) view of the object to be predicted. So one natural question to ask here is, \textit{“obviously, most objects encompass an area of more than one grid cell, and thus would there be a simple mathematical way to assign more of those grid cells try to predict the class and coordinates of the object together with the center grid cell ?”}. Some of the advantages of doing this are (a) reduce imbalance, (b) faster training to converge to the bounding box as now multiple grid cells target the same object at once, (c) increase the chance of predicting tight fit bounding box (d) give grid-based detectors such as YOLOv3 a multi-perspective view  rather than a single point view of the objects. Our multi-grid assignment tries to answers the above question, and we explain it as follow: consider $\left(c'_x,c'_y\right)$  be any grid cell within the distance of $d \in \{-1,0,1\} $ from $\left(c_x,\ c_y\right)$, or mathematically $\left(c'_x,c'_y\right) = \left(c_x + d_x,c_y+d_y\right)$ where $d_x$ and $d_y$ are distance $d$ in x and y directions from the $(c_x, c_y)$ point respectively. Based on the value of d, this equation refers to all the grids marked 0 to 8 in Figure \ref{multigridassignment}(b). Then instead of $\left(c_x,c_y\right)$ based equations of (5-9), we can rewrite a general one using $\left(c'_x,c'_y\right)$ that applies for  any of the grids labeled 0 to 8 as in \Crefrange{equ:nine}{equ:twelve}:

\begin{equation}\label{equ:nine}
	x\ =\ \left(c'_x\ +\ t'_x\right)\times g_w
\end{equation}
\begin{equation}\label{equ:ten}
	y\ =\ \left(c'_y\ +\ t'_y\right)\times g_h
\end{equation}
\begin{equation}\label{equ:eleven}
	b_w\ =\ a_w\times e^{t_w}
\end{equation}
\begin{equation}\label{equ:twelve}
	b_h\ =\ a_h\times e^{t_h}
\end{equation}
Where $t’_x= \mp{d} + t_x$ and  $t’_y= \mp{d} + t_y$. Note that now the bounding box parameter $(t’_x,\ t’_y)$, will have a range of $[-1, 2]$, unlike the $[0, 1]$ range of $(t_x, t_y)$. Figure \ref{objectencoding} shows the ground-truth annotation of the expected output of our object detector. 

\section{Training}
\subsection{The Detection Network: MultiGridDet}
MultiGridDet is an object detection network we repurposed by removing six darknet convolutional blocks from YOLOv3 to make it lighter and faster. A convolutional block has one Conv2D layer followed by a Batch Normalization layer followed by a LeakyRelu layer. The removed blocks are not from the classification backbone, that is, Darknet53. Instead, we removed them from the three multi-scale detection output networks or heads, two from each output network. Though usually deep networks perform well, too deep networks also tend to overfit quickly or drastically reduce the network’s speed. 

In addition to stripping the six convolutional blocks, we also adopt DenseYOLO’s output head. In YOLOv3, each output layer has a tensor shape \textbf{$g_w\times g_h\times k\times\left(5+n\right)$}, where $g_w\times g_h$ is the total grid cells, $k$ is the number of anchors, and $n$ is the total number of object classes. In DenseYOLO, the output layer tensor has a shape \textbf{$g_w\times g_h\times\left(5+k+n\right)$} which means approximately \textbf{$k$} times fewer parameters on the output layer. Moreover, DenseYOLO introduces a novel approach by making an \textit{anchor box a predictable parameter} similar to an object’s class and bounding box prediction. Thus in MultiGridDet, we opt for DenseYOLO’s lighter approach on the output layer. In general, MultiGridDet has less convolutional block and a \textbf{lighter} output head compared to YOLOv3, thus relatively \textbf{faster}.

\subsection{The Loss function}
Like DenseYOLO, our loss function has four parts: class prediction loss, location or coordinate prediction loss, anchor prediction loss, and objectness confidence loss. 

\subsubsection{\textbf{Class prediction loss (error)}}
Our class prediction loss is a simple binary cross-entropy loss calculated over all grid cells labeled to have contained an object of interest. 

\subsubsection{\textbf{Anchor prediction loss (error)}}:- Our anchor prediction loss is also a binary cross-entropy loss in which we train our network to pick an appropriate anchor out of a given set of anchors. We generated nine anchor boxes, 3 for each scale, using IoU (Intersection over Union) based K-means using the same approach as YOLOv3. For example, an anchor with the highest IoU against the ground truth bounding box is assigned to an object during ground truth annotation. And during training, we train the network to learn to pick the anchor that gives the highest IoU to a given object, a concept introduced by DenseYOLO. 

\subsubsection{\textbf{Coordinate prediction loss (error)}}
As shown in Figure \ref{objectencoding}, every object bounding box has four parameters $({t'}_x,\ {t'}_y,t_w,t_h)$ related to the actual bounding box coordinates using \Crefrange{equ:nine}{equ:twelve}.  Accordingly, the network will predict the corresponding bounding box parameters  $(\hat{t}_x,\hat{t}_y,\hat{t}_w,\hat{t}_h)$. As explained in an earlier section, the $({t'}_x,\ {t'}_y)$ corresponds to an object’s center coordinate and has a value in the range $[-1, 2]$. Thus the corresponding network output must pass through an activation function whose output value must also be in the same range. However, the common activation functions such as  tanh have a range $[-1, 1]$, sigmoid $[0, 1]$, and Relu or LeakyRelu have a range either $[0, \infty]$ or $[-\infty, +\infty]$, respectively. We experimented with various custom activations, or simple mapping functions, but finally figured out using tanh and sigmoid activation functions in combination works very well since the output of the sum of the two functions is bounded in the range $[-1, 2]$. 

Let the direct output of the detection network corresponding to $({t'}_x,\ {t'}_y)$ before passing through an activation be $(\hat{z}_x,\hat{z}_y)$. Using \Crefrange{equ:thirteen}{equ:fourteen}, we convert $(\hat{z}_x,\hat{z}_y)$ into $(\hat{t}_x,\hat{t}_y)$. 
\begin{equation}\label{equ:thirteen}
	\hat{t}_x=\tanh(\beta\times\hat{z}_x)\ + \sigma(\beta\times\hat{z}_x)
\end{equation}
\begin{equation}\label{equ:fourteen}
	\hat{t}_y=\tanh(\beta\times\hat{z}_y)\ + \sigma(\beta\times\hat{z}_y)
\end{equation}

\begin{figure*}[h]
\centerline{\includegraphics[width=0.75\textwidth]{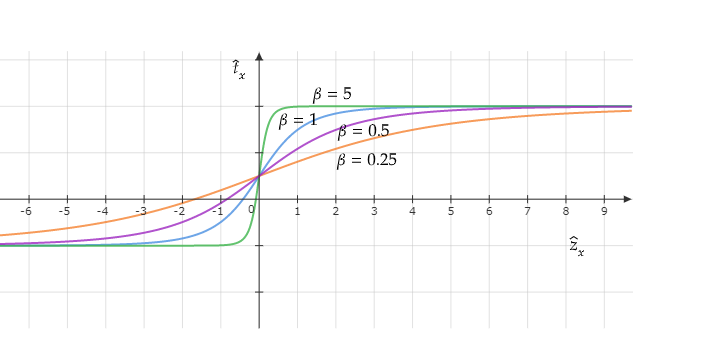}}
\caption{\textbf{Coordinate activation function plot with different $\beta$ values}}
\label{activationplot}
\end{figure*}

As shown in Figure \ref{activationplot}, equation \ref{equ:thirteen} and \ref{equ:fourteen} smoothly transforms the network output $(\hat{z}_x,\hat{ z}_y)$ to the desired output range. The $\beta$ in equations is to horizontally expand the $tanh$ and sigmoid function to prevent quick saturation of these functions. $\beta$ should be picked from range $[0, 1]$ since values above 1 make bounding box prediction unstable during training. In our case we set $\beta=0.25$ and during inference also we use the same value for $\beta$.  

Finally, we calculate coordinate prediction loss using mean square error as given in \cref{equ:fifteen}. 

\begin{equation}
\begin{aligned}\label{equ:fifteen}
    lcrd_{ij}={} & \lambda_{coord}\mathds{1}_{ij}^{obj}\left[\left(\sqrt{x_{ij}}-\sqrt{\hat{x}_{ij}}\right)^2+ \left(\sqrt{y_{ij}}-\sqrt{\hat{y}_{ij}}\right)^2\right]+\\
    & \lambda_{coord}\mathds{1}_{ij}^{obj}\left[\left(\sqrt{b_{wij}}-{\hat{b}}_{wij}\right)^2+\left(\sqrt{b_{hij}}-{\sqrt{\hat{b}}_{hij}}\right)^2\right] 
\end{aligned}
\end{equation}
\begin{equation}
\lambda_{coord} = - \lambda\log(IoU_{score_{ij}})
\end{equation}
\begin{equation}
loss_{coord} = \frac{1}{m}\sum_{i=0}^{g_w}\sum_{j=0}^{g_h}lcrd_{ij}
\end{equation}

$g_w$ and $g_h$ are the total number of grid cells horizontally and vertically, respectively, whereas ${1^{obj}}_{ij}=1$ if the grid cell has an object and otherwise equals zero. $m$ refers to the batch size. ${IoU}_{score_ij}$ has a value of 0 to 1 depending on how much the predicted bounding box overlaps with the ground truth. The logarithmic coefficient we introduced in the coordinate loss plays a significant role. It penalizes incorrect bounding box prediction and rewards the more accurate ones logarithmically, similar to what focal-loss of RetinaNet intends to achieve, except ours is for localization rather than classification.

\subsubsection{\textbf{Objectness confidence loss}}
Fourth part of our loss function evaluates objectness confidence of the predicted bounding box. Our objectness confidence loss  has two parts, as seen in equation 6, one for grids labeled to have contained an object, that is $t_0=1$, and those that are not, meaning $t_0=0$. 
\begin{equation}
\begin{aligned}
loss_{conf}= & \frac{1}{m}\sum_{i=0}^{g_w}\sum_{j=0}^{g_h}\mathds{1}_{ij}^{obj}\times{BCE}_{obj_{ij}} + \\
& \frac{1}{m}\sum_{i=0}^{g_w}\sum_{j=0}^{g_h}\mathds{1}_{ij}^{noobj}\times{BCE}_{noobj}
\end{aligned}
\end{equation}

\begin{equation}
{BCE}_{obj_{ij}}=-\log{\left({\hat{t}}_0\right)}
\end{equation}
\begin{equation}
{BCE}_{noobj_{ij}}=-\log{\left(1-{\hat{y}}_{\left[\cdots,4:\right]}\right)}
\end{equation}
The objectness confidence is similar to YOLOv3 because we both use binary cross-entropy loss, except in our case, the no object loss part tries to make classification, anchor prediction, and objectness confidence to have probability zero.

\subsection{Data Augmentation}  
The other significant contribution of our work is our offline copy-paste-based data augmentation. As much as careful design of artificial intelligence model is essential, a neat and tremendous amount of training and validation data are also mandatory for a better performing network, especially for an object detection network. Recently copy-paste-based augmentation techniques such as simple alpha blending of two or more images MixUp, CutMix and Mosaic augmentations are reported to increase object detectors’ performances. In this work, we implement our own unique and more robust offline copy-paste data augmentation to increase training data significantly. 

In general, our offline copy-paste artificial training image synthesis works as follows: First, we download thousands of background objectless images, meaning images without our object of interest, from google images using a simple image search script using keywords such as landmarks, rain, forest, amusement parks, deserts, cities, wallpapers. We then iteratively pick $p$ number of objects and their bounding boxes from random $q$ images of the entire training dataset. We then generate all possible combinations of the $p$ bounding boxes selected using their index as ID. From the set of the combinations, we pick a subset of bounding boxes that satisfies the following two conditions:
\begin{itemize}
\item if arranged in some random order side by side, they must  fit within a given target background image area 
\item and should efficiently utilize the background image space in its entirety or at least most part of it without the objects overlap.
\end{itemize}

\begin{figure*}[h]
\centerline{\includegraphics[width=1.0\textwidth]{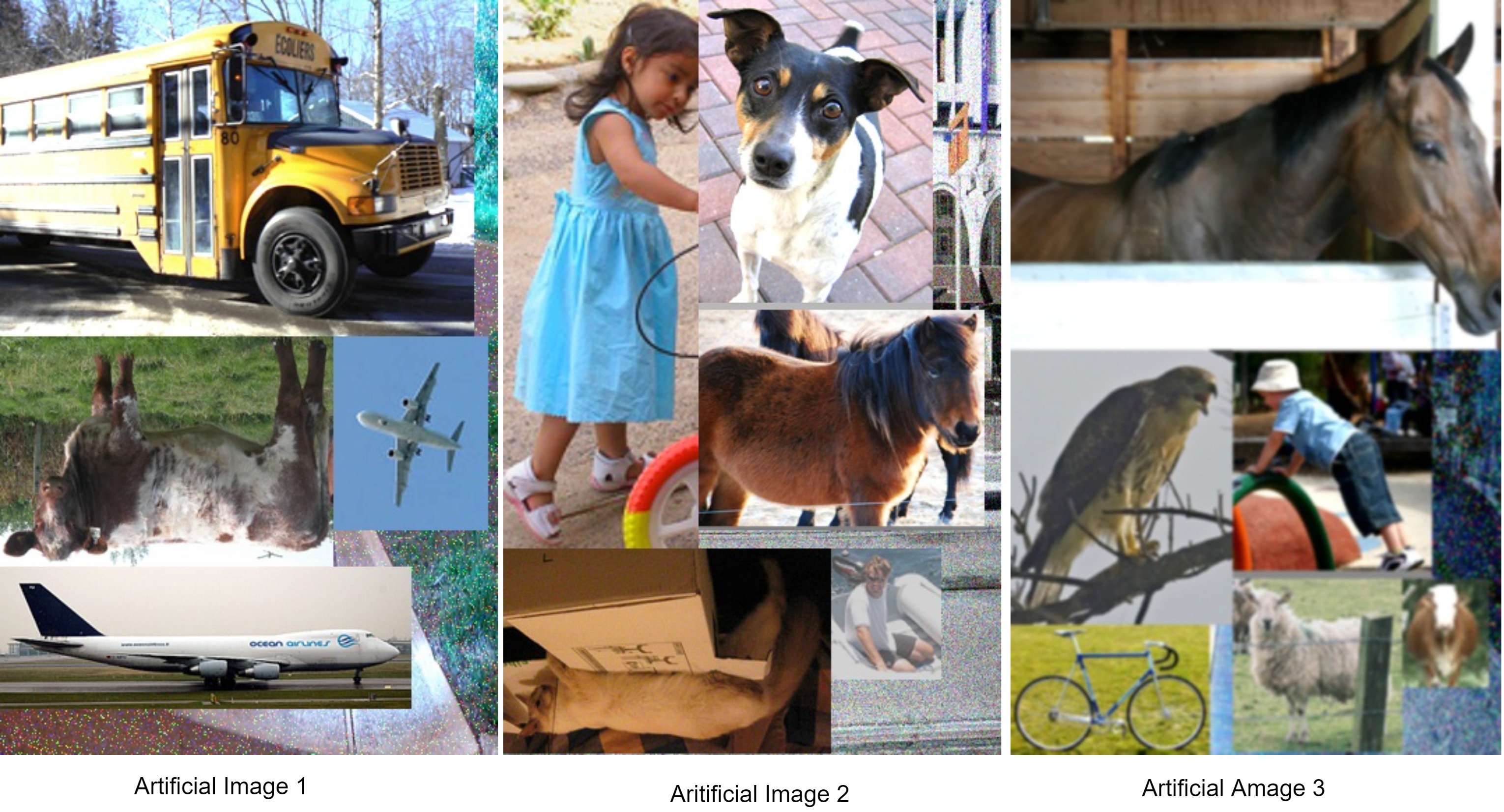}}
\caption{\textbf{Sample Offline Copy-Paste Generated Artificial Images}}
\label{cpimages}
\end{figure*}

Following the above approach, we generate hundreds of thousands of artificial images. Moreover, before copy-pasting an object from one image onto the background, we randomly do various common augmentations on the individual object.  During training, we randomly implement simple and common augmentations to the training minibatch. Fig. \ref{cpimages} shows three sample artificially synthesized images using our offline copy paste augmentation. The figure shows that the artificial images comprise objects that often will not appear together, reassuring the training dataset’s robustness. To prevent the network from learning the copy-paste edges, we add an offset of 10 to 15 pixels in all four sides of the object when copying from the source image, thus assuring the bounding box will not rest on the paste borders. As explained earlier, we also passed each object through one or more common augmentation (flip, brightness, contrast, etc.) before pasting on the background image to prevent early overfitting of the network on the training dataset.

\begin{table}
\begin{center}
\begin{tabular}{c|c|c}
\textbf{Detection Frameworks}&\textbf{mAP} \\
\hline
Fast R-CNN \cite{fasterrcnn} &70  \\
YOLO \cite{yolov1} &63.4  \\
SSD300 \cite{ssd}  &74.3  \\
SSD500 cite{ssd}  &76.8  \\
\hline
YOLOv2 416 \cite{yolov2} &76.8  \\
YOLOv2 544 \cite{yolov2} &78.6  \\
\hline

\textbf{MultiGridDet 416x416 (ours)} &\textbf{83.5}  \\
\end{tabular}
\caption{\textbf{Performance Comparison on Pascal VOC 2007 test set}} 
\label{voc2007}
\end{center}
\end{table}

\section{Experiment}
\begin{table*}[htbp]

\begin{center}
\begin{tabularx}{0.9\textwidth}{@{}l|l|*{20}{C}c@{}}
\hline
Model&data&mAP&aero&bike&bird&boat&bottle&bus&car&cat&chair&cow&table&dog&horse&mbike&person&plant&sheep&sofa&train&tv\\
\hline
Faster \cite{fasterrcnn}&07+12&73.2&76.5&79.0&70.9&65.5&52.1&83.1&84.7&86.4&52.0&81.9&65.7&84.8&84.6&77.5&76.7&38.8&73.6&73.9&83.0&72.6\\
SSD300 \cite{ssd}&07+12&74.3&75.5&80.2&72.3&66.3&47.6&83.0&84.2&86.1&54.7&78.3&73.9&84.5&85.3&82.6&76.2&48.6&73.6&73.9&83.4&74.0\\
SSD512 \cite{ssd}&07+12&76.8&82.4&84.7&78.4&73.8&53.2&86.2&87.5&86.0&57.8&83.1&70.2&84.9&85.2&83.9&79.7&50.3&77.9&73.9&82.5&75.3\\
\hline
\textbf{MultiGridDet}&07+12+CP &\textbf{83.5}&87&93&84&70&73&96&88&93&65&83&80&90&92&92&87&55&85&78&93&84\\
\end{tabularx}
\caption{Individual Pascal VOC 2007 dataset classes mAP. }
\label{individualVoc2007}
\end{center}
\end{table*}

To test our multi-grid redundant object annotation and our offline copy-paste data augmentation, we repurposed the YOLOv3 network into a lighter and faster network, dubbed MultiGridDet, as explained in the earlier section. We perform training on two well-known object detection datasets, namely Pasca-VOC (VOC 2007 + 2012) and COCO datasets. 

We supplement both datasets with artificial images we generated using our offline copy-paste augmentation. We downloaded about 10,000 background images from Google Images using simple search keywords and a script. Then using these background images and total Pascal VOC 2007 and 2012 training and validation images, which constitute about 16K images, we generate an additional 200K artificial images for pascal VOC object detection. In total, we increased the initial 16K Pascal VOC 2007 + 2012 training + validation set images to 216K images and used a validation split of 0.2 so that $80\%$ of the total data are used for training while the remaining $20\%$ are for validation.   

Similarly, we used the same 10K background images and the original 118K COCO images to generate another 200K artificial COCO images. This increases our COCO dataset to 318K images. Similar to the strategy we used on the Pascal VOC dataset, we used a validation split of 0.2 to train object detection on the COCO dataset as well. It is good to note that, though we artificially generated hundreds of thousands of new images, our artificially generated images objects are from the same dataset, picked randomly, individually augmented, and pasted on a randomly picked background image. 

We used the Darknet53 weight file from YOLOv3 authors to train the detector. For the first 50 epochs, we trained only the detection head by freezing the Darknet53 weight file and 150 more epochs by unfreezing the whole network. We start the training with learning rate $1e^{-4}$, and after the $75^{th}$  epochs, we started using cosine decay to update the learning rate. Throughout the training, we used Adam optimizer.  We had access only to 2 Tesla V100 32 GB Nvidia GPU, which limited our training batch size to 64 (32 per GPU) and made the training take longer, restricting us from testing our network performance with other backbones such as ResNet. Next, we will discuss our experiment’s result on the two datasets and compare them with other well-known object detection networks.

\textbf{Pascal VOC 2007 test set}:- Pascal VOC 2007 test set has about 5k test images in 20 class categories. It is one of the widely used generic object detection datasets for comparing the performance of general-purpose object detectors. Accordingly, we tested our MultiGridDet performance using Pascal VOC mean average precision (mAP) metrics at IOU (intersection over union) 0.5. Table \ref{voc2007} shows the mAP performance comparison of MultiGridDet against other state-of-the-art one-stage detectors and equivalent two-stage detectors. As seen from the table, our detector significantly outperforms all older versions of YOLO, YOLOv1, and YOLOv2, including all variants of SSD, RetinaNet, and Fast RCNN. Table \ref{individualVoc2007}  further shows our detector’s per class mAP score in detail. The authors of YOLOv3 never reported the performance of the original YOLOv3 on the Pascal VOC dataset. However, following their training and data augmentation approaches detailed in paper \cite{yolov3}, we retrained YOLOv3 on the combined Pascal VOC 2007 and 2012 training set and achieved a maximum mAP of $77.63 \%$ at input image size $608 \time 608$. This score is much lower than our MultiGridDet score of $83.5\%$ mAP at an input image resolution $416 \time 416$. 

\textbf{MS COCO test set}:- COCO dataset is the most challenging dataset for object detection, typically due to its massively unfair under and over-representation of object class categories and object scale imbalance in the dataset. Nonetheless, it is a more generic dataset consisting of 80 class categories and more robust performance measurement metrics; an mAP averaged over 11 IoU ranges $[0.5 - 1.0]$ referred to as AP(average precision).  Accordingly, we trained our MultiGridDet on the COCO dataset and obtained an AP of $31.8\%$, a little less than YOLOv3’s $33\%$ AP at $608 \times 608$ input image size as shown in Table \ref{cocoAP}. However, on large images, that is objects with a bounding box area above $96^2$ according to COCO metrics, MultiGridDet by far outperforms YOLOv3’s AP $41.9\%$ by $+15.5\%$, scoring AP $57.4\%$. MultiGridDet is poor on small and medium image detections; only $11\% AP$ against YOLOv3’s $18.3\%$ AP on small objects and  $24.6\%$ AP on medium images against YOLOv3 $35.5\%$ AP. Objects such as bottles usually appear in smaller sizes and crowed, whereas objects such as boats and potted plants usually appear in widely irregular shapes and sizes. These are objects MultiGridDet struggled to detect correctly.

\begin{figure*}[h]
\centerline{\includegraphics[width=1.0\textwidth]{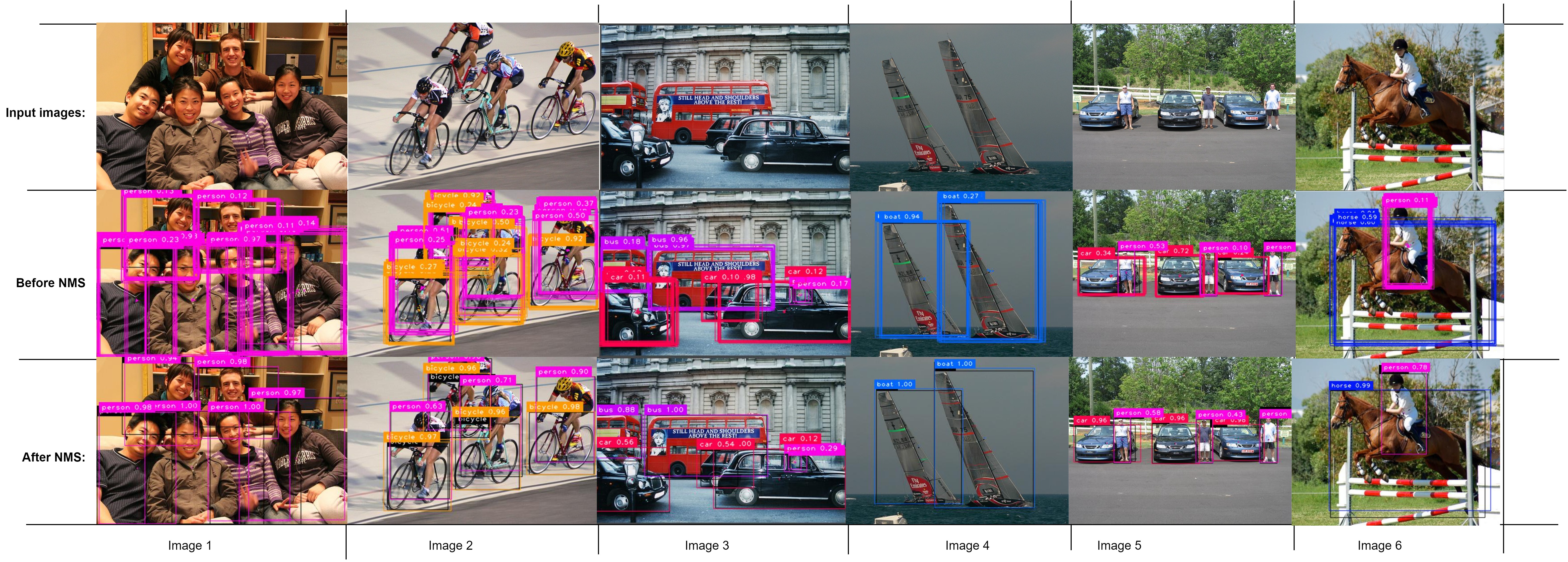}}
\caption{\textbf{Sample MultiGridDet output on randomly selected Pascal VOC 2007 test set images.} As seen from the figure the first row shows six input images, whereas the second row shows the prediction of the network before non-max-suppression (NMS) and the last row shows the final bounding box prediction of MultiGridDet on the input image after NMS thresholding. }
\label{sample_output}
\end{figure*}

Finally, to illustrate the quality of MultiGridDet bounding box prediction, we visualize the prediction of six randomly picked images from the Pascal VOC 2007 test dataset. Fig.  \ref{sample_output} shows these visualization. As seen from the figure, almost in all cases, the predicted unfiltered bounding boxes overlap perfectly, proving tight-fit bounding box prediction.  

In summary, since small objects have tiny widths or height, some even smaller than the $8 \time 8$ grid size, thus they are effectively annotated with a single grid cell, whereas larger objects will have a maximum of 9 grid cells to annotate and predict them. Probably, the addition of more fine-grained output layers, typically an output layer with $2\time 2$ and/or $4\time 4$ grid cell sizes in addition to the $8 \time 8$, $16 \time 16$, and $32 \time 32$ grid cells, might help to increase the performance of MultiGridDet on small object detections. 

We have also compared MultiGridDet inference speed with YOLOv3 on a standard personal laptop with Nvidia GPU Geforce 1060 and 16 GB RAM Intel Core i7-7700HQ CPU 2.80GHz processor. On average, YOLOv3 at $416 \times 416$ input image for 80 class COCO dataset takes 0.149 seconds to infer an image, the overall time spent from reading input image, preprocess, predict, and draw bounding boxes back on the image and display it or save it in a directory. For MultiGridDet at the same input resolution and same dataset, it takes only 0.103 seconds. On video object detection at $416 \times 416$ 80 Class COCO dataset YOLOv3 reaches detection speed of 6FPS (frame per second) whereas MultiGridDet reaches upto 9FPS. Note that the computer we used is not the same as the one the author used, and it has a much slower GPU. In general, in the speed test, MultiGridDet is significantly faster than YOLOv3.

\begin{table}[h]

\begin{center}
\begin{tabular}{l@{\hskip 0.05in}|c@{\hskip 0.05in}c@{\hskip 0.05in}c@{\hskip 0.05in}|c@{\hskip 0.05in}c@{\hskip 0.05in}c@{\hskip 0.05in}}
\setlength{\tabcolsep}{1pt}
\textbf{Models} &\textbf{AP} & \textbf{AP50} &\textbf{AP75} & \textbf{APS} &\textbf{APM} & \textbf{APL} \\
\hline
\textit{Two-stage methods}  &  & & & & &\\
Faster R-CNN+++ \cite{He_2016_CVPR} & 34.9& 55.7& 37.4& 15.6& 38.7& 50.9\\
Faster R-CNN w FPN \cite{lin2017feature} &36.2& 59.1& 39.0& 18.2& 39.0& 48.2\\
Faster R-CNN by G-RMI \cite{huang2017speed} &34.7& 55.5& 36.7& 13.5& 38.1& 52.0\\
Faster R-CNN w TDM \cite{shrivastava2016beyond}  & 36.8& 57.7& 39.2& 16.2& 39.8& 52.1\\
\hline
\textit{One-stage methods}  &  & & & & & \\

YOLOv2 \cite{yolov2} &21.6&44.0&19.2&5.0&22.4&35.5\\
YOLOv3 \cite{yolov3} 608& 33.0& 57.9& 34.4& 18.3& 35.4& 41.9\\
DSSD513 \cite{fu2017dssd} & 33.2& 53.3& 35.2 &13.0 &35.4& 51.1\\
RetinaNet \cite{retinanet} & 40.8& 61.1& 44.1& 24.1 &44.2& 51.2\\
DDGNet \cite{ddgnet} &28.8& 51.2& 30.8& 12.1 &39.4& 50.7\\
DenseYOLO\cite{denseyolo}  &29.03& 50.4& 32.6& 13.11 &33.2& 40.3\\
\hline
\textbf{MultiGridDet @608x608} &31.8& 52.1& 40.7& 11.0 &24.6& 57.4\\
\hline

\end{tabular}
\caption{AP Performance on COCO test set}
\label{cocoAP}
\end{center}
\end{table} 

\section{Conclusion}
In this paper, we proposed a new and alternative object detection implementation for one-stage YOLO-like object detectors that rely on a matrix of grid cells. It is a lightweight, faster, and commendably accurate detector with a prospect for further improvement that addresses the poor performance on the infinitesimally tiny objects of the COCO dataset. A straightforward technique is probably to add finer output scales, for example, $2 \times 2$ or $4 \times 4$, so that the multi-grid annotation could also be implemented on those tiny objects. Another significant contribution we achieved in this work is our unique data augmentation technique that vastly increases object detection training sets without needing additional external dataset. Finally, as future work, we would like to tackle small object detection challenges and try to use MultiGridDet on object tracking and segmentation challenges.

\bibliographystyle{IEEEtran}
\bibliography{ddgnet}

\end{document}